%% file: main.tex
\documentclass[runningheads]{llncs}

\usepackage{graphicx}

\usepackage{amssymb}
\usepackage{bbm}
\usepackage{amsmath}
\usepackage{amsfonts}
\usepackage{mathtools}

\usepackage{dirtytalk}

\usepackage[sectionbib]{natbib}
\usepackage{tikz}
\usepackage{forest}

\usepackage{algpseudocode}
\labelformat{algocf}{Algorithm\,#1}
\usepackage{algorithm}

\usepackage{wrapfig}
\usepackage{adjustbox}

\input{macro}

\usepackage{pdfpages}

\usepackage[misc]{ifsym}

\toctitle
\tocauthor 

\begin{document}

\usetikzlibrary{positioning}
\usetikzlibrary{arrows.meta}

\title{SMACE: A New Method for the Interpretability of Composite Decision Systems
}
\titlerunning{SMACE: Semi-Model-Agnostic Contextual Explainer}

\author{Gianluigi Lopardo\inst{1}[\Letter] \and
        Damien Garreau\inst{1} \and
        Frédéric Precioso\inst{2} \and
        Greger Ottosson\inst{3}
}
\authorrunning{G. Lopardo et al.}
%
\institute{Université Côte d'Azur, Inria, CNRS, LJAD, France \and
           Université Côte d'Azur, Inria, CNRS, I3S, France \and 
           IBM France}
\maketitle              

\begin{abstract}
Interpretability is a pressing issue for decision systems.
Many \emph{post hoc} methods have been proposed to explain the predictions of a single machine learning model.
However, business processes and decision systems are rarely centered around a unique model. 
These systems combine multiple models that produce key predictions, and then apply rules to generate the final decision.
To explain such decisions, we propose the Semi-Model-Agnostic Contextual Explainer (SMACE), a new interpretability method that combines a geometric approach for decision rules with existing interpretability methods for machine learning models to generate an intuitive feature ranking tailored to the end user. 
We show that established model-agnostic approaches produce poor results on tabular data in this setting, in particular giving the same importance to several features, whereas SMACE can rank them in a meaningful way. 

\end{abstract}

\section{Introduction}
Machine learning is increasingly being leveraged in systems that make automated decisions. 
However, the massive adoption of artificial intelligence in many industries is hindered by mistrust, due to the lack of explanations to support specific decisions \citep{jan2020ai}.
Interpretability is deeply linked to trust and, as a result of public concern, has become a regulatory issue. 
For example, the European guidelines for trustworthy AI\footnote{https://digital-strategy.ec.europa.eu/en/library/ethics-guidelines-trustworthy-ai} recommend that \say{AI systems and their decisions should be explained in a manner adapted to the stakeholder concerned.} 

While numerous interpretability methods for single machine learning models exist \citep{linardatos2021explainable}, in many practical applications, a decision is rarely made by a unique model.
In fact, composite AI systems, combining machine learning models together with explicit rules, are very popular, particularly in business settings.
Incorporating decision rules is important, for two main reasons.
Firstly, \textit{decision rules are crucial for expressing policies that can change (even very quickly) over time}. 
For example, depending on last quarter's financial results, a company might be more or less risk-averse and therefore have a more or less conservative policy. 
Using an individual machine learning model would require to retrain it with new data each time the policy changes.
In contrast, with a rule-based system, risk aversion can be managed by changing only a rule.
Secondly, \textit{machine learning models are not suitable for incorporating strict rules}. Indeed, while often a policy may represent a soft preference, in many cases we may have strict rules, due to domain needs or regulation. 
For example, we may have to require that clients' age be over $21$ in order to offer them a service.
It is typically difficult to account for such strict rules in a machine learning setting. 

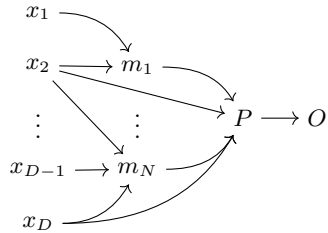
\begin{figure}
    \centering
    \input{tikz/telco_graph}
    \caption{Structure of a composite decision system with $D$ input features $x_1,\ldots,x_D$, and $N$ models $m_1,\ldots,m_N$. 
    A decision policy $P$ (\emph{i.e.}, a set of decision rules) is finally applied to produce an outcome $O$.
    Note that in general both the models and the rules take a subset of input features as input, tough not necessarily the same.}
    \label{fig:telco_graph}
\end{figure}

We focus our study on tabular data, most commonly used in businesses' day-to-day operations, often corresponding to customer records. 
Our interest in this paper is the interpretability of composite decision-making systems that include multiple machine learning models aggregated through decision rules in the form  
\[\texttt{if \{premise\}, then \{consequence\}}.\]
Here, \texttt{premise} is a logical conjunction of conditions on input attributes (\emph{e.g.}, age of a customer) and outputs of machine learning models (\emph{e.g.}, the churn risk of a customer); \texttt{consequence} is a decision concerning a user (\emph{e.g.}, propose a new offer to a customer).
A phone company's policy for proposing a new offer can be
\[ \texttt{if} \ \texttt{age}\leq 45 \ \texttt{and} \ \texttt{churn\_risk} \geq 0.5, \ \texttt{then} \ \texttt{offer} \ 10\% \ \texttt{discount}.\] 
On the one hand, a number of additional challenges arise in this framework (see Section~\ref{sec:challenges}).
On the other hand, there is knowledge we can leverage: we know the decision policy and how the models are aggregated.
It is worth exploiting this information instead of considering the whole system as a black-box and being completely model-agnostic.
In contrast, we want to be agnostic about the nature of individual models: we call this situation \say{semi-model-agnostic.}

In this setting, we present the \textit{Semi-Model-Agnostic Contextual Explainer} (SMACE), a novel interpretability method for composite decision systems that combines a geometric approach (for decision rules) with existing interpretability solutions (for machine learning models) to generate explanations based on feature importance. 
The key idea of SMACE is to agglomerate individual models explanations in a manner similar to that used by the whole decision system.
By making the appropriate assumptions (see Section~\ref{sec:assumptions}), we can see a decision system as a decision tree where some nodes refer to machine learning models. 
In a nutshell, we agglomerate the explanations for each model in a linear fashion, following the structure of this tree.
We therefore combine an \emph{ad hoc} method for the interpretability of decision trees, with \emph{post hoc} methods for the models. 

\paragraph{Contributions.} The main contributions of this paper are
\begin{itemize}
    \item The description of a new method, SMACE, for the interpretability of composite decision-making systems;
    \item The Python implementation of SMACE, available as an open source package at \url{https://github.com/gianluigilopardo/smace};
    \item The evaluation of SMACE \emph{vs} some popular methods showing that the latter perform poorly in our setting.
\end{itemize}
The rest of the paper is organized as follows.
In Section~\ref{sec:related}, we briefly present some related work on both decision trees and \emph{post hoc} methods for machine learning.
Section~\ref{sec:challenges} outlines the main challenges we want to address.
In Section~\ref{sec:smace} the mechanisms behind SMACE are explained step by step; an overview is given in Section~\ref{sec:overview}.
Finally, we provide an evaluation of our method compared to established \emph{post hoc} solutions in Section~\ref{sec:evaluation}, before concluding in Section~\ref{sec:conclusion}.


\section{Related Work}
\label{sec:related}

A decision policy can be embedded in a decision tree. 
Small CART trees \citep{breiman1984classification} are intrinsically interpretable, thanks to their simple structure.
However, as the number of nodes grows, interpretability becomes more challenging. 
\citet{alvarez2004explaining} and \citet{alvarez2009explaining} propose to study the partition generated by the tree in the feature space to rank features by importance.
A similar approach has been used to build interpretable random forests \citep{benard2021sirus}.
We develop a solution inspired by this idea based on the distance between a point and the decision boundaries generated by the tree. 
The main difference in our setting is that each node can be a machine learning model.

Indeed, we also need to deal with machine learning interpretability.
LIME \citep{ribeiro2016should} explains the prediction of any model by locally approximating it with a simpler, intrinsically interpretable linear surrogate.
\citet{upadhyay2021extending} extends LIME to business processes, by modifying the sampling.
Anchors \citep{ribeiro2018anchors} extracts sufficient conditions for a certain prediction, in the form of rules.
SHAP \citep{lundberg2017unified} addresses this problem from a Game Theory perspective, where each input feature is a player, by estimating Shapley values \citep{shapley1953value}.
Despite the solid theoretical foundation, there is concern \citep{kumar2020problems} about its suitability for explanations.
\citet{labreuche2018explaining} leverages Shapley values to explain the result of aggregation models for Multi-Criteria Decision Aiding.
However, their solution requires full knowledge of the models involved, whereas we want to be agnostic about individual models.
SMACE requires feature importance measures, provided for instance by LIME and SHAP (or different approaches as proposed by \citet{framling2022contextual}).  

Overall, perturbation-based methods have some drawbacks and are not always reliable \citep{slack2020fooling}.
In addition, methods using linear surrogates are not suitable to deal with step functions (\emph{e.g.}, the ones encoded by strict decision rules), which often leads to attributing the same contribution to multiple features.
In the case of LIME for tabular data this behavior was pointed out by \citet{garreau2020looking} and \citet{garreau2020explaining}.


\section{Challenges}
\label{sec:challenges} 

As mentioned in the previous section, the field of interpretable machine learning has many unresolved issues.
When trying to explain a decision that relies on multiple machine learning models, a number of additional problems arise:
\begin{itemize}
    \item \textit{Rule-induced nonlinearities}: decision rules will cause sharp borders in the decision space. 
    For example: a car rental rule might state \say{\texttt{age} of renter must be above $21$}. 
    Explanations for a machine learning based risk assessment close to the decision boundary $\texttt{age} = 21$, \emph{e.g.}, must accurately indicate \texttt{age} as an important feature.
    \item \textit{Out-of-distribution sampling}: the decision rules surrounding a machine learning model will eliminate a portion of the decision space.
    Explanatory methods based on sampling like LIME and SHAP are known to distort explanations because of this (see Section \ref{sec:related}).
    \item \textit{Combinations of decision rules and machine learning}: for a specific decision, a subset of rules triggered and a machine learning-based prediction was generated. 
    How do we compose a prediction based on both sources? 
    \item \textit{Multiple machine learning models}: when multiple models are involved in a decision, we must also be able to aggregate multiple feature contributions. 
    These may be (partially) overlapping and conflicting.
\end{itemize}
In addition, we want to have two desirable properties: \textit{(1) the contribution associated with a feature must be positive if it satisfies the condition, negative otherwise}; \textit{(2) the magnitude of the contribution associated with a feature must be greater the closer its value is to the decision boundary}.


\section{SMACE}
\label{sec:smace}

We now present SMACE in more details, starting with a thorough description of our setting in Section~\ref{sec:smace-setting} and a discussion of our assumptions in Section~\ref{sec:assumptions}. 
Section~\ref{sec:overview} contains the overview of the method, with additional details in Section~\ref{sec:explaining-results}, \ref{sec:explain_rules}, and~\ref{sec:overall_explanations}. 


\subsection{Setting}
\label{sec:smace-setting}

Let $x\in\mathbb{R}^{Q\times D}$ be the input data, where each row $x^{(i)}=(x_1,\ldots,x_D)^\top\in\mathbb{R}^D$ is an instance and $D$ is the cardinality of the \textit{input features set} $F$.
Let the set $M=\{m_1,\ldots,m_N\}$ be the set of models.
We will refer to their outputs $m_1(x),\ldots,m_N(x)$ as the \textit{internal features}, whose values we also denote as $y^{(1)},\ldots,y^{(N)}$ when there is no ambiguity.
The union of input and internal features is the set of $D+N$ \textit{features} to which the decision policy can be applied. 

We define $\tilde{x}\coloneqq\left(x_1,\ldots,x_D,m_1(x),\ldots,m_N(x)\right)^\top$ as the completion of $x$ with the outputs of the $N$ models.
Likewise, we call $\xi=\left(\xi_1,\ldots,\xi_D\right)^\top$ the example to be explained and $\tilde{\xi}=\left(\xi_1,\ldots,\xi_D,m_1(\xi),\ldots,m_N(\xi)\right)^\top$ its completion.
A decision rule $R$ is formally defined by a set of conditions on the features in the form $\tilde{x}_j\geq\tau$, for some cutoff $\tau\in\mathbb{R}$.
Figure~\ref{fig:telco_graph} illustrates a generic composite decision system.


\subsection{Assumptions}
\label{sec:assumptions}

The definition of SMACE is based on three assumptions required to frame the setting.
Ideas for solving some of their limitations are discussed in Section~\ref{sec:conclusion}. 
\begin{assumption}\label{assump:one}
    Decision rules only refer to numerical values.
\end{assumption}
This assumption allows us to take a simple geometric approach for the explainability of the decision tree. 
Note that this does not imply any restriction on the input of the machine learning models, that can still be categorical.
\begin{assumption}\label{assump:two}
    Each decision rule is related to a single feature, without taking into account feature interactions. 
\end{assumption}
For instance, this assumption excludes conditions like $\texttt{if} \; \tilde{x}_1\geq\tilde{x}_2$. 
Geometrically, this implies decision trees with splits parallel to the axes, such as CART \citep{breiman1984classification}, C4.5 \citep{quinlan1993c45}, and ID3 \citep{quinlan1986induction}. 
\begin{assumption}\label{assump:three}
    The machine learning models only use input features to make predictions.
\end{assumption}
We disregard the cases in which a machine learning model takes as input the output of other machine learning models. 
We remark that this is a very reasonable assumption that covers most real-world applications.
Note that assumptions \ref{assump:one} and \ref{assump:two} refer to the decision rules, while Assumption \ref{assump:three} is the only referring to the machine learning models and does not concern their nature. 


\subsection{Overview}
\label{sec:overview}
For each example $\xi$ whose decision we want to explain, we first perform two parallel steps:
\begin{itemize}
    \item \textbf{Explain the results of the models}: for each machine learning model $m$, we derive the (normalized) contribution $\hat{\phi}^{(m)}_j$ for each input feature $j$.
    By default, SMACE relies on KernelSHAP to allocate these importance values;
    \item \textbf{Explain the rule-based decision}: measure the contribution $r_j$ of each feature (that is, each input feature and each internal feature directly involved in the decision policy), through Algorithm \ref{alg:rules}. 
\end{itemize}
Then, to get the \textbf{overall explanations} (see Algorithm \ref{alg:smace}), we combine these partial explanations.
The total contribution of the input feature $j\in F$ to the decision for a given instance is
\begin{equation}\label{eq:aggregate}
        e_j = r_j + \sum_{m\in M} r_m\hat{\phi}^{(m)}_j \,.
\end{equation}
That is, we weight the contribution of input features to each model with the contribution of that model in the decision rule, and we add the direct contribution of feature $j$ to the decision rule (if a feature is not directly involved in a decision rule, its contribution is zero).


\subsection{Explaining the results of the models}
\label{sec:explaining-results}
We need to attribute the output of each machine learning model to its input values.
For instance, this is what KernelSHAP does, and by default SMACE relies on it. 
In any case, SMACE requires a measure of feature importance for the input features, but not necessarily based on SHAP. 
Any other measure of feature importance is possible. 
Given the contribution $\phi_j^{(m)}$ of each input feature $j$ for each machine learning model $m$ we define the normalized contribution as
\begin{equation}\label{eq:modelcontribution}
    \hat{\phi}^{(m)}_j=
    \begin{cases}
        \frac{\abs{\phi^{(m)}_j}}{\displaystyle\sum_{i\in F}\abs{\phi_i^{(m)}}}\,, & \quad \text{if}\:\displaystyle\max_{i\in F}{\abs{\phi_i^{(m)}}}\neq 0 \,, \\
        0\,, & \quad \text{otherwise}.    
    \end{cases} 
\end{equation} 

Indeed, two models $m_k$ and $m_h$ might give results $y^{(k)}$ and $y^{(h)}$ on very different scales, for instance because they do not have the same unit. 
In the example above, we may have models computing the churn risk and the life time value. 
The first value estimates a probability, so it belongs to $[0,1]$, while the second is the expected economic return that the company may get from a customer, and it could be a quantity scaling as thousands of euros. 
In general, if $m_k$ predicts the churn risk and $m_h$ predicts the life time value, for a feature $j$ in input to both models, we might expect $\abs{\phi_j^{(h)}}\gg\abs{\phi_j^{(k)}}$.
In order to have a meaningful comparison between the models, we therefore need to scale the $\phi$ values and we use as scale factor the sum of the $\phi$ values for each model.
The quantities $\hat{\phi}$ defined by means of Eq.~\eqref{eq:modelcontribution} are of the same order of magnitude and dimensionless, so can be aggregated.
In addition, $\hat{\phi}$ is defined such that 
\[
\forall j\in F\,,\:\forall m\in M\,,\quad 0\leq\hat{\phi}_j^{(m)}\leq 1
\,.
\]
Note that the second part of Eq.~\eqref{eq:modelcontribution} is equivalent to taking the convention $\frac{0}{0}=0$\,: the denominator is zero if and only if each contribution is zero.
The definition implies that if the model $m$ relies on a single feature $j$, the latter will have
\[
\hat{\phi}^{(m)}_j= 1\implies r_m\hat{\phi}^{(m)}_j=r_m\,,
\]
\emph{i.e.}, the whole contribution of the model $m$ to the decision is attributed to the input feature $j$, which in fact is the only one responsible for its output.


\subsection{Explaining the rule-based decision}
\label{sec:explain_rules} 

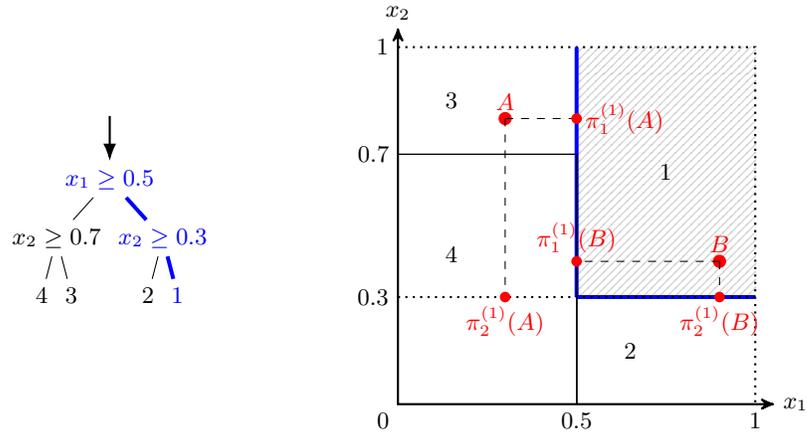
\begin{figure}
    \input{tikz/boundaries}
    \caption{On the left, a decision tree classifier based on $x_1$ and $x_2$.
    In \color{blue}\textbf{blue and bold}\color{black}, the trace for leaf $1$. 
    On the right, the partition it generates.
    $A$ and $B$ are instance points, classified respectively as $3$ and $1$.
    The decision surface for leaf $1$ is in \color{blue} \textbf{blue and bold}\color{black}.
    The dashed lines indicate the distance between the points and the decision boundaries.}
    \label{fig:boundaries}
\end{figure}

In Section~\ref{sec:related} we stated that the set of conditions used by a decision system can be interpreted as a CART tree, such as the one in Figure~\ref{fig:boundaries}, where each split represents a condition on a feature.
A first approach to explain the decision of such a tree can be to show the trace followed by the point within the tree to the user.
However, the trace does not contain enough information to understand the situation: a large change in some conditions may have no impact on the result, whereas a very small increase in one value may lead to a completely different classification, if we are close to a split value. 

In addition, there may be many conditions within a decision rule, and simply listing them all would make it difficult to understand the decision.
In fact, each condition is a split in the decision tree and each split produces a decision boundary. 
The collection of decision boundaries generated by the tree induces a partition of the input space and we call decision surface the union of the boundaries of the different areas corresponding to the different classes.
Because of Assumption \ref{assump:two}, at each point $z\in S$, the decision surface is piecewise-affine, consisting of a list of hyperplanes, each referring to one feature.
By projecting an example point $\tilde{x}$ onto each component $j$ of the surface $S$, we obtain the point $\pi^{(S)}_j(\tilde{x})$ (see Eq. \eqref{eq:proj}) at minimum distance that satisfies the condition on the $j$-th feature (see Figure \ref{fig:boundaries}).
This distance is a measure of the robustness of the decision with respect to changes along feature $j$.
Conversely, the smaller the distance, the more \textit{sensitive} the decision. 

As mentioned in Section \ref{sec:challenges}, we want the method to assign a greater contribution to features with higher sensitivity.
In this way, values close to the decision boundary are highlighted to the end user and the domain expert, who will be able to draw the appropriate conclusions.
The explainability problem is therefore addressed by studying the decision surfaces generated by the decision tree. 

However, to properly compare these contributions, we must first normalize the features. 
We must then query the models on the training set in order to obtain the values $y^{(1)},\ldots,y^{(N)}$.
We thus apply a min-max normalization on both input features
\[
\forall \, i\in\{1,\ldots, Q\} \,,\quad x'^{(i)}_j = \frac{x^{(i)}_j-\min{x_j}}{\max{x_j}-\min{x_j}} \, ,
\]
and internal features, likewise.
In this way, the values of each feature is scaled in $[0,1]$.
For the sake of convenience, we continue to denote the features $x'_i$ and $y'^{(k)}$ as $x_i$ and $y^{(k)}$, but from now on we consider them as scaled.

\input{algo/smace}
\input{algo/rules}

Each decision surface $S$ has as many components (hyperplanes) as there are features defining it.
For instance, the decision surface for leaf $1$ of Figure~\ref{fig:boundaries} has two components: $h_1$ and $h_2$, along $x_1$ and $x_2$, respectively.
The projection $\pi_j^{(S)}(x)$ of point $x$ onto $h_j$ is 
\begin{equation}\label{eq:proj}
    \pi_j^{(S)}(\tilde{x})\in\argmin_{z\in h_j}{\norm{\tilde{x}-z}_2}\,.
\end{equation}
For instance, let us consider the decision tree of Figure \ref{fig:boundaries} and the partition it generates.
Let us say we are interested in leaf $1$ (the grid subspace shown in Figure \ref{fig:boundaries}) generated by the trace in blue.
Example $B$ satisfies both conditions, while $A$ only satisfies the condition on $x_2$.
We also note that the decision for $B$ is very sensitive with respect to changes along axis $x_2$, while it is more robust with respect to $x_1$.
We compute the contribution $r_j$ of a feature $j$ for the classification of point $\tilde{x}$ in leaf $\ell$ by means of Algorithm \ref{alg:rules} as
\begin{equation}\label{eq:contribution}
    r_j(\tilde{x}) = \begin{cases}
    \abs{\tilde{x}_j-\pi_j^{(\ell)}(\tilde{x})}-1\,, & \quad \text{if}\ \tilde{x}_j<h_j \,,  \\
    1-\abs{\tilde{x}_j-\pi_j^{(\ell)}(\tilde{x})}\,, & \quad \text{if}\ \tilde{x}_j\geq h_j \,.
    \end{cases}
\end{equation}
We can see that for point $A$, the feature $x_1$ has a high negative contribution, since it does not satisfy the condition on it, while $x_2$ has a positive contribution.
Point $B$ satisfies both conditions: both features have positive contributions, but $r_2(B)>r_1(B)$, since the decision is more sensitive with respect to $x_2$.  


\subsection{Overall explanations}
\label{sec:overall_explanations}

Finally, once the partial explanations have been obtained, we agglomerate them via Eq. \eqref{eq:aggregate}.
We thus obtain a measure of the importance of features for a specific decision made by a system combining rules and machine learning models.
Our measure of importance highlights the most critical features, those therefore most involved in the decision.
In this way, a domain expert can analyse a decision by focusing on these features to make her or his own qualitative assessment.

\paragraph{Computational cost.}
The most computationally expensive step of SMACE is to get explanations for the underlying models.
It basically consists in $N$ calls to the explainer on (at most) $D$ input features. 
For instance, in the case of KernelSHAP, this would be $N \times 1000 \times D$. 


\section{Evaluation}\label{sec:evaluation}

What makes interpretability even more challenging is the lack of adequate metrics to appropriately assess the quality of explanations.
In this section we compare the results obtained with SMACE and those obtained by applying the default implementations of SHAP\footnote{\url{https://github.com/slundberg/shap}} and LIME\footnote{\url{https://github.com/marcotcr/lime}} on the whole decision system.
We first perform a qualitative analysis on simple use cases, where we can get a complete understanding of the decision provided by the system.
We show that SHAP and LIME do not satisfy the properties stated in Section \ref{sec:explain_rules} and we therefore argue that they are not suitable in this context. 
Finally, we perform a sanity check on aggregate explanations on three different realistic use cases.


\subsection{Qualitative analysis}\label{sec:qualitative}
The input data consists of $1000$ instances, each with three randomly generated components as uniform in $[0,1]^3$. 
Note that decision rules on these data generate partitions analogous to those in Figure \ref{fig:boundaries}, but in dimension $3$. 

\subsubsection{Rules only}\label{sec:eval_rules}
Let us first evaluate the case of a decision system consisting of only three simple conditions applied to only three input features.
The decision policy contains rule $R_1$:
\[\texttt{if} \ x_1\leq 0.5 \ \texttt{and} \ x_2\geq 0.6 \ \texttt{and} \ x_3\geq 0.2 \ \texttt{then} \ 1, \ \texttt{else} \ 0.\]
Note that there are no models, $R_1$ is based solely on the input data.
The method then reduces to the application of Eq. \eqref{eq:contribution}, discussed in Section \ref{sec:explain_rules}.

\paragraph{Example with two violated attributes.} 
Take the example to be explained in an arbitrary position with respect to the boundaries: $\xi^{(1)}=(0.6, 0.1, 0.4)^\top$. 
The decision is $0$, since the rule $R_1$ is not satisfied, indeed the conditions $\xi^{(1)}_1\leq0.5$ and $\xi^{(1)}_2\geq0.6$ are violated.
We want to know why $\xi^{(1)}$ is not classified as $1$ and the contributions of the three features to that decision.
The comparison is shown in Table \ref{tab:eval1_random}.
The results of SMACE are computed (Eq. \eqref{eq:contribution}) as
\[
\begin{cases}
    r_1 = \abs{0.6-0.5}-1 = -0.9\,, \\
    r_2 = \abs{0.1-0.6}-1 = -0.5\,, \\
    r_3 = 1-\abs{0.4-0.2} = 0.8\,.
\end{cases}
\]
\input{eval/rulesonly/random1}
In this case, we see that all the three methods agree in their signs, satisfying property \textit{(1)}.
However, SHAP and LIME attribute the same contribution to $x_1$ and $x_2$ even though the sensitivities of the values are different. 
They do not satisfy property \textit{(2)}: the contribution of $x_1$ should be higher in magnitude than that of $x_2$, since it is closer to the boundary.
This behavior is due to the nonlinearities brought by the decision rules, as mentioned in Section~\ref{sec:related}.
The point is that the sampling is performed in a space away from the boundary, and so by perturbing the example in a small neighborhood, the output does not change.

\paragraph{Slight violation on one attribute.} 
We now consider the specific case where two features are exactly on the decision boundary, while one condition is slightly violated.
Let us consider the example $\xi^{(2)}=(0.51, 0.6, 0.2)^\top$. 
The decision-making system classifies $\xi^{(2)}$ as~$0$ for a slight violation of the rule on the first attribute.
\input{eval/rulesonly/viol1}
In Table \ref{tab:eval1_viol1} we see that SMACE highlights the slight violation of the rule on $x_1$.


\subsubsection{Simple hybrid system}

Let us add two simple linear models $m_1$ and $m_2$. 
The models are defined as
\[
\begin{cases}
    m_1(x) = 1 x_2 + 2 x_3\,, \\
    m_2(x) = 700 x_1 -  500 x_2 + 1000 x_3\,.
\end{cases}
\]
We are interested in rule $R_3$:
\[\texttt{if}\: x_1\leq0.5\: \texttt{and}\: x_2\geq0.6\: \texttt{and}\: m_1\geq 1\: \texttt{and}\: m_2\leq 600 \: \texttt{then}\: 1\,,\: \texttt{else}\: 0\,,\]
and we want to explain the decision for $\xi^{(1)}\,$.
The comparison on the whole system is in Table~\ref{tab:paper}.
Again, LIME and SHAP are producing identical results on $x_1$ and $x_2$, missing useful information.
SMACE disagrees with the other methods on the sign of $x_3$, correctly giving a negative sign (Property \textit{(1)}).
Indeed, the input feature $x_3$ has a high contribution for the model $m_2$ and $m_2$ is not satisfying the condition ($m_2(\xi^{(1)})=770>600$), so it has a negative contribution.

\input{eval/linearmodels/case_paper}

\smallskip

By analyzing individual explanations, we have shown that SMACE produces meaningful results by assigning each feature a contribution proportional to its distance from the boundary.
On the contrary, SHAP and LIME often assign the same contribution to different features, not providing useful information about the relative importance of each feature. 


\subsection{Sanity check}\label{sec:sanity}

In the previous section, we showed that SMACE is able to produce meaningful feature attributions. 
We now demonstrate that SMACE also retains an ability to identify the set of features contributing negatively to a decision, regardless of individual attribution. 
If a feature contributes negatively, it means it must be moved to meet its condition. 
Correctly identifying negative features is a desirable property: to change the decision, each of them must be moved. 

We consider $100$ random instances which do not satisfy the rules (described in the supplementary), from three different datasets, and we apply SMACE, SHAP, and LIME. 
For each method, we extract the set of negative features. 
Note that to be sure that the rule will be satisfied, each negative feature should be shifted to a specific value: none of the three methods is giving this information. 
We then generate $1000$ samples by shifting negative features with a local perturbation. 
The average decision made on these perturbed samples is an indicator of the quality of the explanations provided by each of the three methods.

\begin{figure}[t!]
    \centering
    \includegraphics[width=0.45\textwidth]{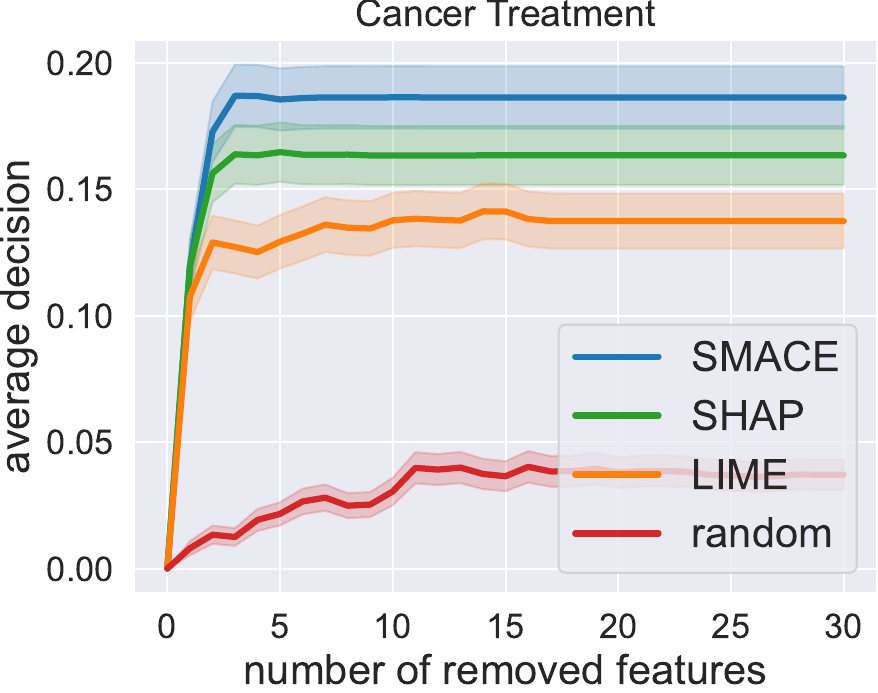}
    \includegraphics[width=0.45\textwidth]{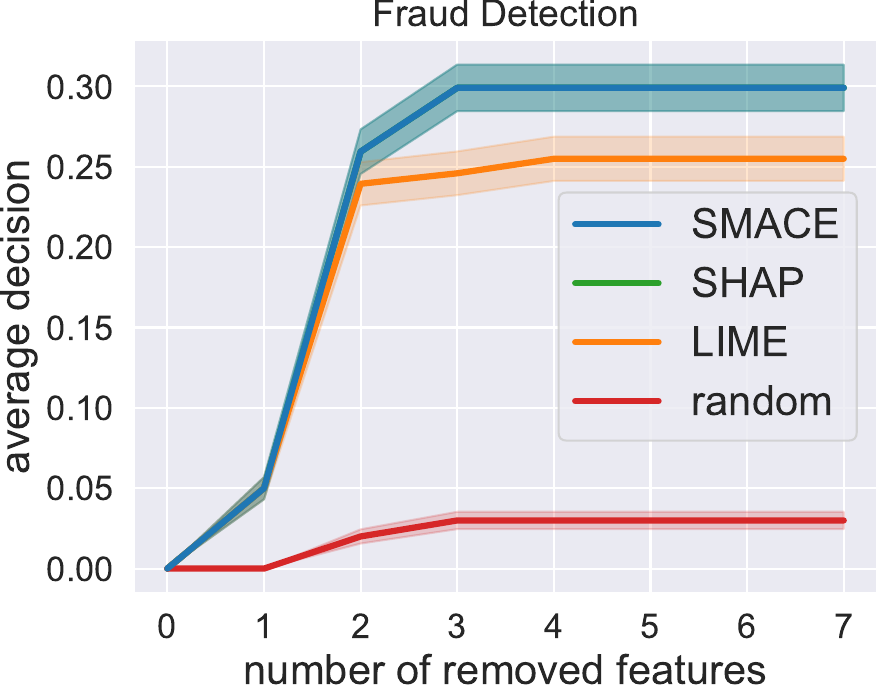}
    \includegraphics[width=0.45\textwidth]{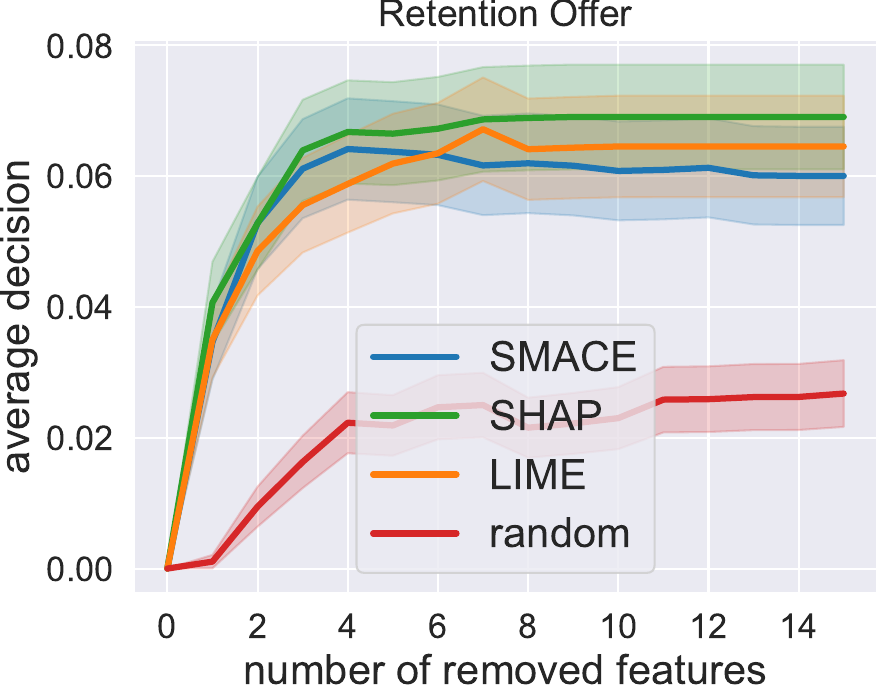}
    \caption{Comparison of SMACE, SHAP, and LIME on the ability to identify the set of features contributing negatively to a decision, regardless of individual attribution. 
    Correctly identifying negative features is a desirable property: to change the decision, each of them must be moved.
    When the conditions are not met, the three methods are used to extract the negative features, and we generate perturbed samples around the original values. 
    We then compare the average decision made on the samples. 
    }
    \label{fig:comparison}
\end{figure}

\textbf{Cancer treatment}
A machine learning model is trained to predict whether a breast cancer is benign or malignant from information about its size and structure. 
An automated decision system is then applied to decide on treatment: if the risk of the tumor being malignant is too high, it proceeds in full reliance on the model. 
If, on the other hand, the probability is low, but the size and composition of the tumor are suspicious, further investigation is carried out. 
The decision system consists of $30$ continuous \textit{input features} and $1$ \textit{internal feature} (coming from the model).
We use the \textit{Breast Cancer Wisconsin Data Set}.\footnote{\url{ https://www.kaggle.com/datasets/uciml/breast-cancer-wisconsin-data}}

In this example, we want to explain \emph{why} the treatment was not proposed, \emph{i.e.}, which input features are negatively contributing to the decision. 
Given the large number of parameters to be analyzed, it is useful to order them by importance, in order to speed up the investigation by giving the right priorities. 
The graph at the top left of the Figure \ref{fig:comparison} shows the comparison. 
SMACE curve is always above the others: it is better at detecting negative features. 

\smallskip

\textbf{Fraud Detection}
A financial authority must track mobile money transactions, promptly halting anomalous transactions suspected of fraud. 
The authority uses a decision-making system to approve or block transactions, according to a \emph{fraud score}, computed through a machine learning classifier, and the amount and balanced involved in the transaction.
We use the \textit{Synthetic Financial Datasets For Fraud Detection}\footnote{\url{ https://www.kaggle.com/ealaxi/paysim1}}
As before, we extract and perturb the negative features set for each method. 

The graph at the top right of Figure \ref{fig:comparison} shows that SMACE and SHAP are on par. 
In this decision system, the conditions based on the input features matter significantly more than the one on the model. 
This means that SMACE explanations are almost entirely based on Eq. \eqref{eq:contribution} and, consistently with what we saw in Section \ref{sec:eval_rules}, SMACE and SHAP are able to extract the correct set of negative features. 
However, we remark that SHAP is likely to assign them the same (negative) contribution: SMACE carries more information. 

\smallskip

\textbf{Retention Offer}
Let us consider a mobile phone company which wants to predict if a customer is going to leave for a competitor, and to decide if a retention offer should be made, while not spending more on retention than the value of retaining the customer.
The decision policy is based on information about the customer and their subscription (input features), and two models (producing internal features) predicting the \emph{churn risk} (\emph{i.e.}, the likelihood that the customer will cancel their subscription) and the \emph{lifetime value} (\emph{i.e.}, the expected revenue generated by the customer if retained).
We use the IBM \textit{Telco Churn} dataset.\footnote{\url{ https://github.com/IBMDataScience/DSX-DemoCenter/tree/master/DSX-Local-Telco-Churn-master}}

In this example, we want to explain \emph{why} a retention offer was not made, in terms of the original input features. 
In practice, the features that are contributing negatively should be moved to meet the conditions. 
Note that this use case is characterized by the presence of many categorical input features (see Assumption~\ref{assump:one}): this is a stress test for SMACE. 
Figure~\ref{fig:comparison} shows that SMACE is comparable with the state of the art in extracting the right set of negative features: error bars are overlapping. 
However, it is only a partial measure of quality, since the ranking of features is ignored.
As seen in Section~\ref{sec:qualitative}, SMACE is also able to rank these features by sensitivity.

\smallskip

We compared the ability of SMACE, SHAP, and LIME to extract features that are negatively contributing to a decision and should therefore be moved to change it. 
SMACE is best when applied to the standard context: one or more models and several continuous features (Cancer Treatment). 
SHAP tends to extract the same set of negative features as SMACE when the impact of models is absent or insignificant (Fraud Detection). 
SMACE loses performance when many categorical features are involved in the decision: however, the error bars of the three methods are overlapping (Retention Offer). 
In addition, as seen in Section~\ref{sec:qualitative}, SMACE is also able to rank these features by sensitivity, while SHAP and LIME tend to attribute identical explanations.


\section{Conclusion and Future Work}\label{sec:conclusion}

We addressed the problem of explaining decisions produced by a decision-making system composed of both machine learning models and decision rules.
We proposed SMACE, to generate feature importance based explanations. 
Up to the best of our knowledge, it is the first method specifically designed for these systems. 
SMACE approaches the problem with a projection-based solution to explain the rule-based decision and by aggregating it with models explanations.
We finally showed that model-agnostic approaches designed to explain machine learning models are not well-suited for this problem, due to the complications coming with the rules.
In contrast, SMACE provides meaningful results by meeting our requirements, \emph{i.e.}, adapting to the needs of the end user.

In future work, we plan to extend SMACE, making it usable in a wider range of applications.
A particularly interesting approach to include categorical features in the rules is implemented in CatBoost \citep{prokhorenkova2018catboost}, a gradient boosting toolkit. 
The idea is to group categories by \textit{target statistics}, which can replace them.
SMACE could also be generalized to more complex model configurations, where some models take as input the output of other models.
One natural extension would be to recursively weight the importance of each model with the contribution it brings for other models.

\subsubsection{Acknowledgments.}
This work has been supported by the French government, through the NIM-ML project (ANR-21-CE23-0005-01) and through the 3IA Côte d'Azur (ANR-19-P3IA-0002), and by EU Horizon 2020 project AI4Media (contract no. 951911, \url{https://ai4media.eu/}).




\section*{Supplementary material}

# Supplementary material for the article
# SMACE: A New Method for the Interpretability of Composite Decision Systems

Gianluigi Lopardo, Damien Garreau, Frédéric Precioso, Greger Ottosson

In this supplementary material, we provide additional information about the experiments. In Section 1 the structure of SMACE code is briefly described. In Section 2.3 we present the composite decision system used in the paper as *Realistic use case*, detailing the data, the machine learning models and the decision policy. In Section 3 we illustrate the experiment setting.

## 1 Code description

The code is stored in two main directories: `smace` and `evaluation`. The first one contains the code behind the method. In the second one there is the evaluation. Some simple examples are given as Jupyter Notebook[1] in the `notebooks` folder. Aggregated experiments of the type described in Section 3 are instead contained in the `experiments` folder and the results saved in the `results` subfolder.

## 2 Realistic use cases

### 2.1 Cancer treatment

This use case is based on the dataset *Breast Cancer Wisconsin (Diagnostic) Data Set* avaiable on Kaggle.[2] It contains 30 continuous features, computed from a digitized image of a breast mass. They describe characteristics of the cell nuclei present in the image. A Logistic Regression model is trained to classify breast cancer in malignant or benign.

---

[1] https://jupyter.org/
[2] https://www.kaggle.com/datasets/uciml/breast-cancer-wisconsin-data

The decision policy consists of the rule

```
if malignant_risk ≤ 0.6 and radius_se ≥ 0.2 and area_mean ≥ 300
   and perimeter_mean ≥ 50 and concavity_mean ≥ 0.05
   and compactness_worst ≥ 0.2
   then 1, else 0.
```

### 2.2 Fraud detection

This use case is based on thefraud dataset *Synthetic Financial Datasets For Fraud Detection* avaiable on Kaggle.[3] It contains financial logs from a mobile money service. We remark data the dataset is strongly unbalanced: less than $1\%$ of the instances are actually flagged as Fraud. An XGBoost Classifier is trained to classify a transaction as a fraud.

The decision policy consists of the rule

```
if fraud_risk ≤ 0.01 and amount ≥ 1000 and newbalanceOrig ≥ 1000
   and oldbalanceOrg ≥ 10000 and newbalanceDest ≥ 1000
   and oldbalanceDest ≥ 1000
   then 1, else 0.
```

### 2.3 Retention Offer

In this experiment, we use the churn dataset *DSX Local Telco Churn demo* used by IBM in demo product.[4] It contains information about the customers of a telephone company. It consists of 1799 instances, 15 input features (see Table 1) and two target variables: CHURN and LTV. The first is the *churn risk* of a customer, *i.e.*, the likelihood that the customer will cancel their subscription, and the second is the *lifetime value*, *i.e.*, the expected revenue generated by the customer if retained. Note that categorical features are present in the dataset. Recalling what was expressed in Section 4, we cannot address the case where categorical variables are directly present in the decision policy, but they do not pose a problem when used as input to machine learning models.

We train a XGBoost Classifier to predict CHURN and a XGBoost Regressor to predict LTV. XGBoost is a scalable gradient boosting system presented by Chen and Guestrin [2016]. It stands for "eXtreme Gradient Boosting." It is a scalable, fast and well-performing tree boosting system, widely used in the data science community. It is designed for optimizing computational resources as it performs different optimization improvements which make it better than other boosting technique.

We remark that both models are trained on the whole input features. Note that as CHURN we use the churn risk likelihood obtained via the `predict_proba` function of XGBoost. This is a value is a value in $[0, 1]$ and since default threshold is set to $0.5$, the condition CHURN $\geq 0.5$ (respectively, CHURN $< 0.5$) equals CHURN $= 1$ (respectively, CHURN $= 0$).

---

[3]https://www.kaggle.com/ealaxi/paysim1
[4]https://github.com/IBMDataScience/DSX-DemoCenter/tree/master/DSX-Local-Telco-Churn-master

Table 1: Customer input features for Retention offer use case.

| feature | type |
|---|---|
| Gender | boolean |
| Status | categorical |
| Children | boolean |
| Est. Income | numerical |
| Car Owner | numerical |
| Age | numerical |
| LongDistance | numerical |
| International | numerical |
| Local | numerical |
| Dropped | boolean |
| Paymethod | categorical |
| LocalBilltype | categorical |
| LongDistanceBilltype | categorical |
| Usage | numerical |
| RatePlan | categorical |

The decision policy applied by the company to propose a specific retention offer is

`if Age` $\leq 50$ `and LTV` $\geq 500$ `and CHURN` $\geq 0.5$ `and Usage` $\geq 50$ `and Local` $\leq 200$
`then` $1$, `else` $0$.

## 3 Experiment setting

We want to evaluate the ability of SMACE to detect and possibly rank features that contribute negatively to the decision. The experiment is designed as follows. We select the 100 instances $\xi^{(1)}, \ldots, \xi^{(100)}$ from the dataset that are closer (in norm 2) to the decision boundary and do not satisfy the decision rule. Then, for each instance we apply SMACE, SHAP, and LIME and extract only the features that have a negative contribution. Negative features are then "removed" one-by-one in order of importance. To simulate the removal, we sample 1000 truncated Gaussians $z_1, \ldots, z_{1000}$ as follows. Let $j$ be a negative feature for one example $\xi^{(k)}$, $k = 1, \ldots, 100$. The choice of such a distribution is motivated by the fact that we are dealing with local interpretability and thus using a uniform distribution, for example, would distort the results. Also, some values (such as churn risk in the example above) may be limited in a certain domain.

The truncated Gaussian has parameters

$$b_j = \min_{k \in \{1,\ldots,Q\}} x_{k,j}, B_j = \max_{k \in \{1,\ldots,Q\}} x_{k,j}, \mu_j = \xi_j^{(k)},$$

and

$$\sigma_j^2 = \sqrt{\frac{1}{Q} \sum_{k=1}^{Q} (x_{k,j} - \bar{x}_{k,j})^2}.$$

3

More precisely, $\forall i \in \{1\ldots,1000\}$ $z_{i,j}$ has a density given by

$$\rho_j(t) = \frac{1}{\sigma_j\sqrt{2\pi}} \cdot \frac{\exp\frac{-(t-\mu_j)^2}{2\sigma_j^2}}{\Phi(r_j) - \Phi(\ell_j)} \mathbb{1}\left(t \in [b_j, B_j]\right) , \tag{1}$$

where we set $\ell_j = \frac{b_j - \mu_j}{\sigma_j}$ and $r_j = \frac{B_j - \mu_j}{\sigma_j}$, and $\Phi$ is the cumulative distribution function of a standard Gaussian random variable.

We then count the proportion of this sample that meet all the conditions and average this value for the 100 examples.



\end{document}

%% file: macro.tex
\DeclarePairedDelimiter{\norm}{\lVert}{\rVert} 
\newcommand{\abs}[1]{\left\lvert#1\right\rvert}  

\DeclareMathOperator*{\argmin}{arg\min}
\newtheorem{assumption}{Assumption}

%% file: tikz/telco_graph.tex
\usetikzlibrary{positioning}
\centering
\begin{tikzpicture}
    \node (x1) {$x_1$};
    \node (x2) [below of=x1
    , yshift=+0.3cm] {$x_2$};
    \node (dots) [below of=x2, yshift=+0.3cm] {$\vdots$};
    \node (xD1) [below of=dots, yshift=+0.3cm] {$x_{D-1}$};
    \node (xD) [below of=xD1, yshift=+0.3cm] {$x_D$};
    \node (mdots) [right=of dots, xshift=+0.0cm] {\vdots};
    \node (m1) [above=of mdots, yshift=-0.9cm] {$m_1$};
    \node (mN) [below=of mdots, yshift=+0.9cm] {$m_N$};
    \node (rule) [right=of mdots] {$P$};
    \node (decision) [right=of rule, xshift=-0.5cm] {$O$};
    \draw[->] (x1) to[out=0,in=+120] (m1);
    \draw[->] (x2) to (m1);
    \draw[->] (x2) to (mN);
    \draw[->] (xD1) to  (mN);
    \draw[->] (xD) to[out=0,in=-120] (mN);
    \draw[->] (x2) to (rule);
    \draw[->] (xD) to[out=0,in=-120] (rule);
    \draw[->] (m1) to[out=0,in=+120] (rule);
    \draw[->] (mN) to[out=0,in=-120] (rule);
    \draw[->] (rule) to (decision);
\end{tikzpicture}

%% file: tikz/boundaries.tex
\usetikzlibrary{trees}
\usetikzlibrary{arrows,calc,patterns}

\tikzset{
>=stealth',
help lines/.style={dashed, thick},
axis/.style={<->},
boundary1/.style={thick, line width=1.5pt, color=blue},
boundary/.style={thick, line width=0.5pt},
projection/.style={thick, dotted},
}
\forestset{
    trace/.style={edge+={color=blue,line width=1.5pt}},
}
\begin{tikzpicture}
\begin{forest}
[$x_1\geq 0.5$, trace, color=blue, tikz={\draw[{Latex}-, thick] (.north) --++ (0,0.6);}
    [$x_2\geq 0.7$
        [4]
        [3]    
    ]
    [$x_2\geq 0.3$, trace, color=blue
        [2]   
        [1, trace, color=blue] 
    ]   
] 
\end{forest}
\end{tikzpicture}\hfill
\begin{tikzpicture}[scale=1]
    \coordinate (y) at (0,5);
    \coordinate (x) at (5,0);
    \draw[axis] (y) -- (0,0) --  (x);
    \draw [<->,thick] (0,5) node (yaxis) [above] {$x_2$}
    |- (5,0) node (xaxis) [right] {$x_1$};
    \coordinate (zero) at (0,0); 
    \draw (zero) node [below left] {$0$};
    \coordinate (x2) at ($1*0.95*(y)$); 
    \coordinate (x2_cutoff1) at ($0.7*0.95*(y)$); 
    \coordinate (x2_cutoff2) at ($0.3*0.95*(y)$); 
    \coordinate (x1) at ($1*0.95*(x)$); 
    \coordinate (x1_cutoff) at ($0.5*0.95*(x)$); 
    \coordinate (x2_) at ($0.3*0.95*(x)$); 
    \coordinate (bl) at (0.5*0.95*5, 0.3*5*0.95);
    \coordinate (br) at (1*0.95*5, 0.3*5*0.95);
    \coordinate (al) at (0.5*0.95*5, 1*5*0.95);
    \coordinate (ar) at (1*0.95*5, 1*5*0.95);

    \fill[pattern color=gray, pattern=north east lines, fill opacity=0.4] ($(bl)$) -- ($(al)$) -- ($(ar)$) -- ($(br)$);
    \draw[projection] let \p1=(x2), \p2=(x1) in (\p1) node[left] {$1$} -| (\x2, \y1) -| (\p2) node[below] {$1$};
    \draw[boundary] let \p1=(x2), \p2=(x1_cutoff) in (\p2)  -| (\x2, \y1) -| (\p2) node[below] {$0.5$};
    \draw[boundary1] let \p1=(x2_), \p2=(x2), \p3=(x1_cutoff) in (\x3,\x1) -| (\x3,\y2);
    \draw[boundary] let \p1=(x2_cutoff1), \p2=(x1_cutoff) in (\p1) node[left] {$0.7$} -| (\x2, \y1) -| (\p2);
    \draw[boundary1] let \p1=(x2_cutoff2), \p2=(x1_cutoff), \p3=(x1) in (\x2, \y1) -| (\x3, \y1);
    \draw[projection] let \p1=(x2_cutoff2), \p2=(x1_cutoff), \p3=(x1) in (\p1) node[left] {$0.3$} -| (\x2, \y1) -| (\x3, \y1);

    \draw let \p1=($(x1_cutoff)$), \p2=($(x2_cutoff1)$), \p3=($(x1)-(x1_cutoff)$), \p4=($(x2)-(x2_cutoff1)$) in ($(0.3*\x1, \y2+0.5*\y4)$) node {$3$};
    \draw let \p1=($(x1_cutoff)$), \p2=($(x2_cutoff2)$), \p3=($(x1)-(x1_cutoff)$), \p4=($(x2)-(x2_cutoff2)$) in ($(\x1+0.5*\x3, \y2+0.5*\y4)$) node {$1$};
    \draw let \p1=($(x1_cutoff)$), \p2=($(x2_cutoff1)$), \p3=($(x1)-(x1_cutoff)$), \p4=($(x2)-(x2_cutoff1)$) in ($(0.3*\x1, 0.6*\y2)$) node {$4$};
    \draw let \p1=($(x1_cutoff)$), \p2=($(x2_cutoff2)$), \p3=($(x1)-(x1_cutoff)$), \p4=($(x2)-(x2_cutoff2)$) in ($(\x1+0.3*\x3, 0.5*\y2)$) node {$2$};
    
    \coordinate (A_1) at (0.5*0.95*5, 0.8*5*0.95);
    \coordinate (A_2) at (0.3*0.95*5, 0.3*5*0.95);
    \coordinate (B_1) at (0.5*0.95*5, 0.4*5*0.95);
    \coordinate (B_2) at (0.9*5*0.95, 0.3*5*0.95) ;

    \coordinate (A) at (0.3*5*0.95, 0.8*5*0.95);
    \fill[red] (A) circle (2.5pt);
    \draw (A) node [above, text=red] {$A$};
    \draw[dashed] (A) -- (A_1); \draw[dashed] (A) -- (A_2); 
    \fill[red] (A_1) circle (2pt);
    \fill[red] (A_2) circle (2pt);
    \draw (A_1) node [right, text=red] {$\pi_1^{(1)}(A)$};
    \draw (A_2) node [below, text=red] {$\pi_2^{(1)}(A)$};
    
    \coordinate (B) at (0.9*5*0.95, 0.4*5*0.95);
    \fill[red] (B) circle (2.5pt);
    \draw (B) node [above, text=red] {$B$};
    \draw[dashed] (B) -- (B_1); \draw[dashed] (B) -- (B_2);
    \fill[red] (B_1) circle (2pt);
    \fill[red] (B_2) circle (2pt);
    \draw (B_1) node [above, text=red] {$\pi_1^{(1)}(B)$};
    \draw (B_2) node [below, text=red] {$\pi_2^{(1)}(B)$};
    
\end{tikzpicture}

%% file: algo/smace.tex
\begin{algorithm}
    \caption{Overview of \texttt{smace}.}\label{alg:smace}
    \begin{algorithmic}
        \Function{smace\_explain}{rule $R$ (set of conditions), list of models $M$, example to explain $\xi\in\mathbb{R^D}$}
            \State $\tilde{\xi}\gets\xi\,,\: \phi\gets\{0\}^N\,,\: r\gets\{0\}^{D+N}\,,\: e\gets\{0\}^D\:$\;
            \For{$m\in M$}
                \State $\hat{\phi}^{(m)}\gets\textsc{explain\_model}(\xi,m)$ \Comment{explain the result of model $m$ (Section \ref{sec:explaining-results})}
                \State $\tilde{\xi}\gets\left(\xi_1,\ldots,\xi_D,\ldots,m(\xi)\right)$ 
            \EndFor
            \For{$j=1,\ldots,D+N$}
                \State    $r_j\gets\textsc{rule\_contribution}(R,j,\tilde{\xi})$ \Comment{explain the rule-based decision}
           \EndFor
            \For{$j=1,\ldots,D$}
                \State $e_j\gets r_j + \displaystyle\sum_{m\in M} r_m\hat{\phi}^{(m)}_j$ \Comment{aggregate}
            \EndFor
        \State \Return $e$  
        \EndFunction
    \end{algorithmic}
\end{algorithm}

%% file: algo/rules.tex
\begin{algorithm}
    \caption{Computing \textsc{rule\_contribution}.}\label{alg:rules}
    \begin{algorithmic}
        \Function{rule\_contribution}{rule $R\,,$ variable $j\,,$ example to explain $\tilde{\xi}$}
            \State $S\gets R$  \Comment{projection to the decision surface $S$ generated by $R$}
            \State $\pi_j^{(S)}(\tilde{\xi})\gets\displaystyle\argmin_{z\in h_j}{\norm{\tilde{\xi}-z}_2}$ 
            \If{$\tilde{\xi}$ satisfies condition on $j$}
                \State $r_j\gets 1-\abs{\tilde{\xi}_j-\pi_j^{(S)}(\tilde{\xi})}$
            \Else 
                \State $r_j\gets \abs{\tilde{\xi}_j-\pi_j^{(S)}(\tilde{\xi})}-1$ 
            \EndIf
        \State \Return $r_j$  
        \EndFunction
    \end{algorithmic}
\end{algorithm}

%% file: eval/rulesonly/random1.tex
\begin{table}\caption{Example in generic position, three conditions on three input features.
LIME and SHAP are producing flat explanations on the variables $x_1$ and $x_2$, even if their sensitivities for the decision are very different.
SMACE captures this information.}
    \label{tab:eval1_random}
    \centering
    \begin{tabular}{c c | r r r}
    condition & example $(\xi^{(1)})$ & \textbf{SMACE} & SHAP & LIME \\
    \hline
    $x_1\leq0.5$ & $0.6$ & $-0.9$ & $-0.08$ & $-0.21$ \\
    $x_2\geq0.6$ & $0.1$ & $-0.5$ & $-0.08$ & $-0.21$ \\
    $x_3\geq0.2$ & $0.4$ & $0.8$ & $0.02$ & $0.04$ \\
    \end{tabular}
\end{table}

%% file: eval/rulesonly/viol1.tex
\begin{table}\caption{Slight violation on one attribute, conditions on three input features.
LIME and SHAP do not highlight the high sensitivities for $x_2$ and $x_3$, which are exactly on their respective decision boundary.}
    \label{tab:eval1_viol1}
    \centering
    \begin{tabular}{c c | r r r}
    condition & example $(\xi^{(2)})$ & \textbf{SMACE} & SHAP & LIME \\
    \hline
    $x_1\leq0.5$ & $0.51$ & $-0.99$ & $-0.29$ & $-0.22$ \\
    $x_2\geq0.6$ & $0.60$ & $1.00$ & $0.12$ & $0.14$ \\
    $x_3\geq0.2$ & $0.20$ & $1.00$ & $0.03$ & $-0.20$ \\
    \end{tabular}
\end{table}

%% file: eval/linearmodels/case_paper.tex
\begin{table}
\caption{\label{tab:paper}Simple hybrid system, comparison on the whole decision system.
LIME and SHAP both produce the same explanations for features~$1$ and~$2$.
}
    \centering
    \begin{tabular}{c | r r r}
     example $(\xi^{(1)})$ & \textbf{SMACE} & SHAP & LIME \\
    \hline
    $\xi_1^{(1)}= 0.6$ & $-1.03$ & $-0.08$ & $-0.19$ \\
    $\xi_2^{(1)} = 0.1$ & $-1.73$ & $-0.08$ & $-0.19$ \\
    $\xi_3^{(1)} = 0.4$ & $-0.54$ & $0.02$ & $0.09$ \\
    \end{tabular}
\end{table}